\documentclass[a4paper,12pt]{article}

\usepackage[a4paper,width=165mm]{geometry}
\usepackage[utf8]{inputenc}
\usepackage{epsfig}
\usepackage{subfigure}
\usepackage{calc}
\usepackage{caption}
\usepackage{amssymb}
\usepackage{amstext}
\usepackage{amsmath}
\usepackage{pslatex}
\usepackage{hyperref}

\usepackage{xspace}
\newcommand{\ie}{i.\,e.\xspace}
\newcommand{\eg}{e.\,g.\xspace}

\usepackage{pstricks,pstricks-add}
\def\dpi{6.2831853071795862}
\def\rtw{1.4142135623730951}
\newcommand{\graph}[4][13.4]{
  \psset{unit=0.8}
  \begin{pspicture}(0,#2)(#1,#3)
  \psplot[plotpoints=1000,algebraic=true,linewidth=1pt]{0.001}{#1}{#4}
  \psaxes{->}(0,0)(0,#2)(#1,#3)
  \end{pspicture}}

\newtheorem{scenario}{Scenario}
\newtheorem{definition}{Definition}

\begin{document}

\title{\bf Neural Networks and Continuous Time\thanks{This paper is an extended version of \cite{SR09a}.}}
\author{Frieder Stolzenburg\thanks{e-mail address: \textsf{fstolzenburg@hs-harz.de}}
	\hspace{0pt} and Florian Ruh\thanks{new address: \textsf{f.ruh@etamax.de}}}
\date{Harz University of Applied Sciences\\
	Department of Automation and Computer Sciences\\[\smallskipamount]
	38855~Wernigerode, Germany}

\maketitle

\begin{abstract}
The fields of neural computation and artificial neural networks have developed
much in the last decades. Most of the works in these fields focus on
implementing and/or learning discrete functions or behavior. However, technical,
physical, and also cognitive processes evolve continuously in time. This cannot
be described directly with standard architectures of artificial neural networks
such as multi-layer feed-forward perceptrons. Therefore, in this paper, we will
argue that neural networks modeling continuous time are needed explicitly for
this purpose, because with them the synthesis and analysis of continuous and
possibly periodic processes in time are possible (\eg for robot behavior)
besides computing discrete classification functions (\eg for logical reasoning).
We will relate possible neural network architectures with (hybrid)
automata models that allow to express continuous processes.

\end{abstract}

\section{Introduction}

Cognitive science can be defined as the interdisciplinary study of mind and
intelligence \cite{Tha08}, \ie, how information is represented and
transformed in the brain. Natural or artificial cognitive systems are able to
make decisions, draw conclusions, and classify objects. All these just mentioned
tasks can be implemented more or less adequately by complex logical circuits. Clearly,
the human brain is far more than that. Human beings can learn new knowledge and
behavior. They can interact socially, assigning emotions and intentions to each
other. Therefore, cognitive systems interact with their environment and with
other agents. In consequence, in order to solve complex cognitive tasks, agents
cannot be understood as isolated thinking entities in this context, but as
agents with some kind of body, acting in a (physical) environment. This requires
also the modeling of (continuous) time. Hence, following the lines of cognitive
artificial intelligence research, we will present in this paper a computer
model, based on neural networks, that addresses the above-mentioned issues, such
that cognitive tasks can be simulated~-- including behavioral and even emotional
aspects (\eg music perception).

During the last decades, the field of (artificial) \emph{neural networks} has drawn more
and more attention due to the progress in software engineering with artificial
intelligence. Neural networks have been applied successfully \eg to speech
recognition, image analysis, and in order to construct software agents or
autonomous robots. A basic model in the field is a multi-layer feed-forward
perceptron. It can be automatically trained to solve complex 
classification and other tasks, \eg by the well-known backpropagation algorithm
(cf.~\cite{Hay94,RN10}). Implementing and/or learning discrete functions or
behavior is in the focus of neural networks research.

Nevertheless, technical, physical, and also cognitive processes evolve
continuously in time, especially if several agents are involved. In general,
modeling multiagent systems means to cope with constraints that evolve
according to the continuous dynamics of the environment. This is often
simulated by the use of discrete time steps. In the literature, \emph{hybrid
automata} are considered for the description of such so-called hybrid systems by a mathematical model,
where computational processes interact with physical processes. Their behavior
consists of discrete state transitions plus continuous evolution \cite{Hen96}.
Hybrid automata have been successfully applied especially to technical and
embedded systems, \eg for describing multi-robot behavior \cite{FM+08a,MFS10,RS08}.
However, a feasible procedure for learning hybrid automata
does not seem to be available.

In this paper, we will at first introduce application scenarios that
include complex cognitive, technical, or physical processes for the synthesis 
and analysis of continuous and possibly periodic systems of agent behavior
(Sect.~\ref{sec:scen}). After that, we briefly discuss some related works on neural
networks and hybrid automata wrt. their applicability to timely continuous
systems (Sect.~\ref{sec:works}). Then, we present an enhanced model of
neural networks with continuous time, which we call \emph{continuous-time neural network}
(CTNN) (Sect.~\ref{sec:cnn}), that can simulate the behavior of hybrid
automata as a system that interprets periodic, continuous input and the
response to that. It can also be used for periodicity detection, as needed \eg
in speech or musical cognition, which may be associated with emotions. Finally,
we will end up with conclusions (Sect.~\ref{sec:conc}).

The enhanced neural network model presented here provides not only an adequate
remedy for modeling continuous processes which occur in realistic environments,
but it comes also closer to a more adequate model of the human brain which also
works time-dependent. Hence, the proposed neural network model can contribute to
both engineering-oriented and cognitive artificial intelligence.

\section{Scenarios of Agents in a Continuously Evolving Environment}
\label{sec:scen}

\begin{figure}
	\centering
	\includegraphics[width=0.9\textwidth]{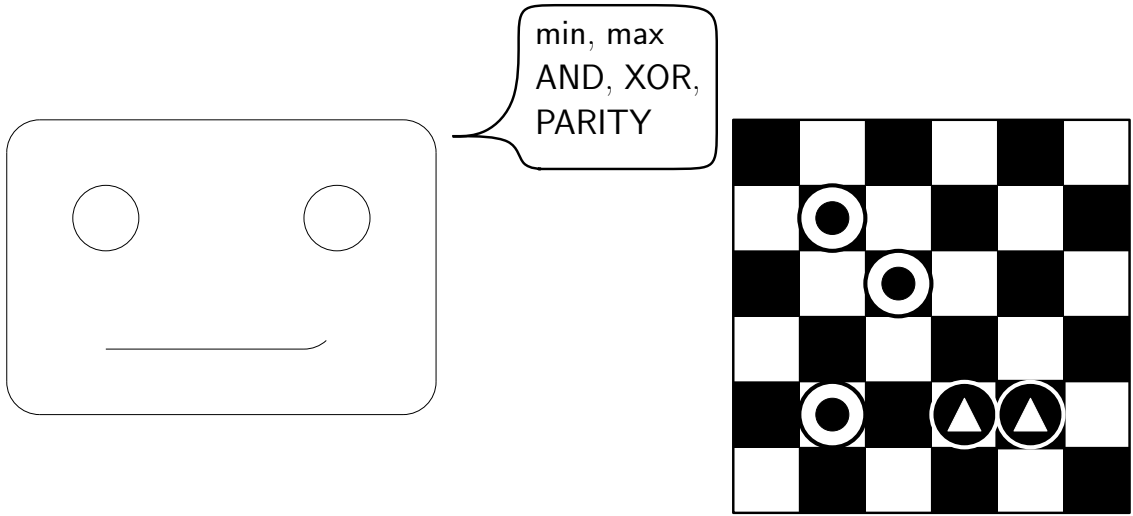}
	\caption{Agent reasoning deductively.}\label{fig:chess}
\end{figure}

\begin{scenario}[deductive reasoning]\label{scen:logic}
Classification tasks like \eg image recognition or playing board games (see
Fig.~\ref{fig:chess}) require deductive reasoning and cognition. In this scenario,
the environment is discrete (according to the classification in \cite{RN10}),
because there is only a limited number of distinct percepts and actions. In
particular, it is not dynamic, \ie, the environment does not change over time,
while the agent is deliberating.
\end{scenario}

Ordinary artificial neural networks allow to solve classification tasks and to
express logical Boolean functions for deductive reasoning directly, \ie functions of the
form $f : X \to Y$, where $X =
(x_1,\dots,x_n)$ represents the input values and $Y = (y_1,\dots,y_m)$ the
output values. Therefore, deductive reasoning can be adequately implemented by
using them. Neural networks in general consist of an interconnected group
of nodes, called \emph{units}, which are programming constructs roughly mimicking the
properties of biological neurons. Standard neural networks such as multi-layer
feed-forward perceptrons have a restricted architecture. There, we have only three types of 
units: input, hidden, and output units, which are connected only in this order
and organized in layers \cite{Hay94,RN10}. While there are always exactly one
input and one output layer, there may be zero, one, or more hidden layers.
It is well-known \cite{Hay94} that every
continuous function that maps intervals of real numbers to some output interval
of real numbers can be approximated arbitrarily closely by a multi-layer
perceptron with just one hidden layer, if we have sigmoidal activation functions,
\ie bounded, non-linear, and monotonously increasing functions, \eg the logistic
function or the hyperbolic tangent ($\tanh$). Multi-layer networks use a variety of
learning techniques, the most popular being backpropagation. In general,
any declarative logical operation can be learned by such a network. However,
many real cognitive or physical processes depend on time, as in the following
scenario. 

\begin{figure}
	\centering
	\includegraphics[width=0.85\textwidth]{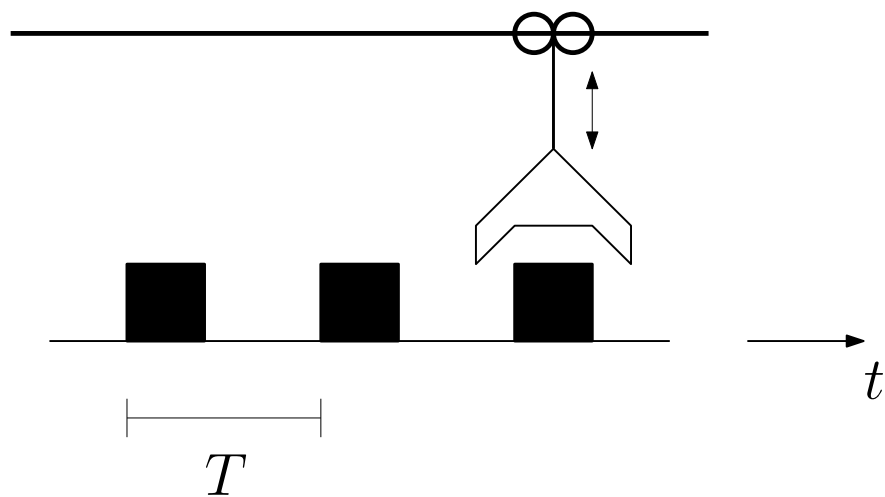}
	\caption{An example robot arm, picking boxes on a conveyor belt.}
	\label{fig:rob}
\end{figure}

\begin{figure}
	\centering
	\includegraphics[width=0.95\textwidth]{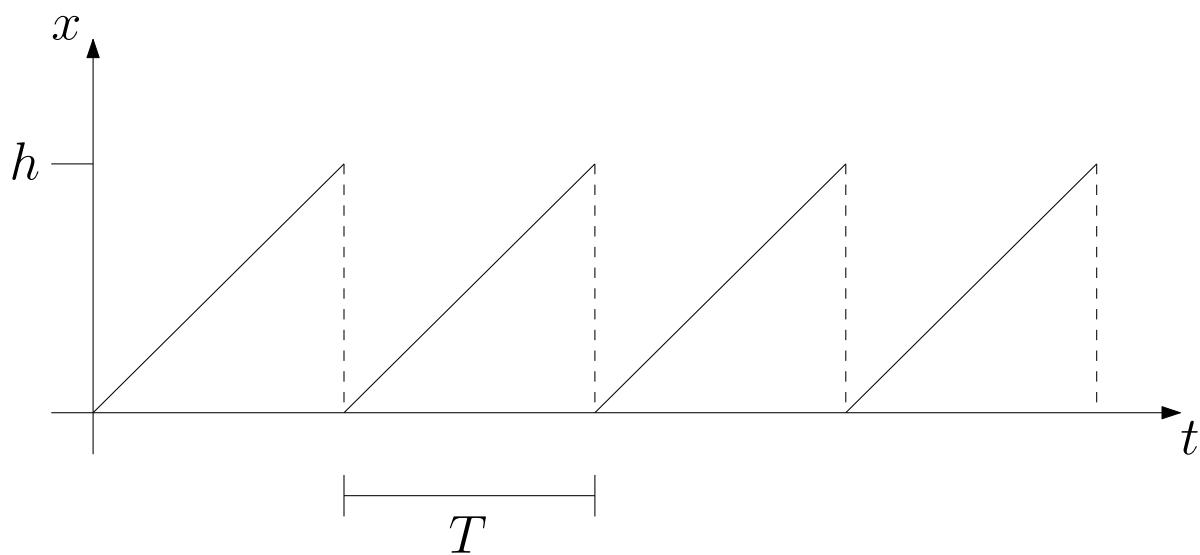}
	\caption{The sawtooth function for the height of the robot arm, assuming
	that it can lower the arm in zero time.}
	\label{fig:saw}
\end{figure}

\begin{scenario}[robot at a conveyor belt]\label{scen:robot}
Let us consider a robot that has to perform a specific routine again and again,
\eg grabbing a brick from a conveyor belt (see Fig.~\ref{fig:rob}).
For the ease of presentation, we abstract from releasing the box, moving
the arm down and grabbing the next one here. In addition, we assume, that the
agent knows the duration $T$ of each episode. For this, Fig.~\ref{fig:saw} shows
the height $h$ of the robot arm depending on the time $t$.
\end{scenario}

This scenario requires the solution of several tasks. In particular, continuous behavior
of the robot agent must be producible for grabbing the bricks continuously and
periodically. Clearly, for synthesis and also for analysis of processes or
behavior, modeling the time $t$ explicitly is necessary, because we have to
model mappings of the form $X(t) \mapsto Y(t)$. For Scenario~\ref{scen:robot},
we assume that the robot has to move its arm up and down within a fixed time
interval $T$. This leads to a sawtooth function, if we consider the dependency
from time (see Fig.~\ref{fig:saw}). Such behavior can be expressed easily by an
automaton model, especially hybrid automata \cite{Hen96} (see Sect.~\ref{sec:ha}).
However, the procedure with hybrid automata mainly is a knowledge-based approach.
Hybrid automata cannot be learned easily by examples as \eg neural networks.

While clearly Scenario~\ref{scen:logic} can be specified directly with ordinary
neural networks, Scenario~\ref{scen:robot} requires to model the time $t$
somehow. This can be achieved by discretizing time, \ie by considering input
values at different discrete time points, $t,\,t-1,\,\dots,\,t-T$ for some time
horizon $T$. Then, we may use $x_i(t),\,x_i(t-1),\,\dots,\,x_i(t-T)$ with $1 \le
i \le n$ as input values. But this procedure has several disadvantages: It
increases the number of input units significantly, namely from only $n$ to
$(T+1) \cdot n$. In addition, it is not clear in this case, what granularity and
past horizon of discrete time should be used.

\begin{figure}
	\centering
	\includegraphics[width=0.9\textwidth]{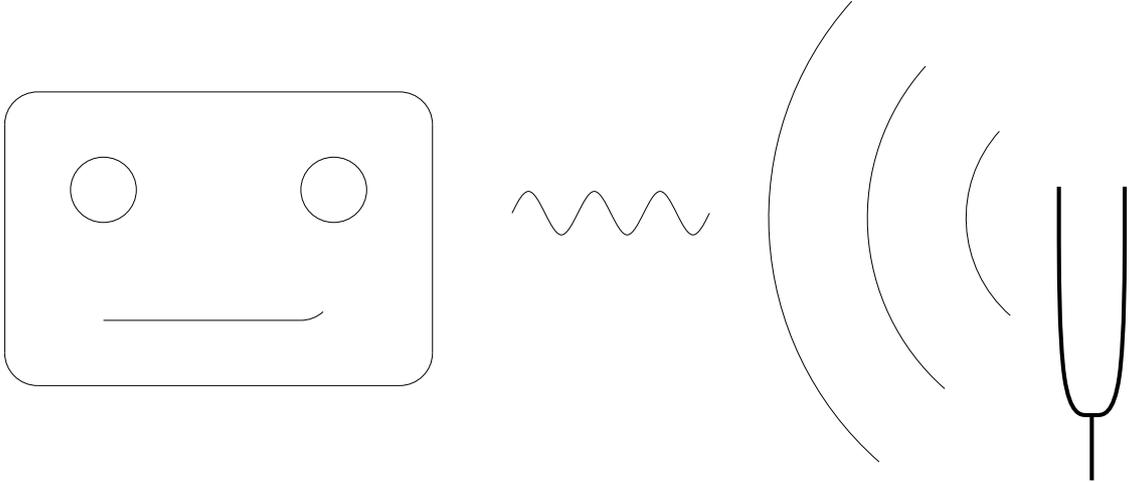}
	\caption{Agent analyzing periodic episodes in the environment.}\label{fig:music}
\end{figure}

Therefore, a presentation by (enhanced) neural networks seems to be a good idea,
that makes use of the (continuous) time $t$ as additional parameter, at least
implicitly. In this context, oscillating periodic behavior must be producible, even
if the input $X$ remains static, \ie constant. For instance, once switching on a robot,
\ie change one input unit from $0$ to $1$, the periodic behavior should hold on,
until the input unit is switched off again (cf.~\cite{MB+08}). Therefore, we
will introduce units, that oscillate, \ie,
whose input may be a fixed value, but whose output yields a sinusoid (see
Sect.~\ref{sec:cnn}, Def.~\ref{def:unit}). By this, we can express periodic
behavior in time by neural networks. Furthermore, we should be able to analyze
behavior and to detect period lengths, which we formulate now:

\begin{scenario}[behavior and periodicity analysis]\label{scen:period}
Before an agent is able to behave adequately in a dynamic environment. For the
example from above (Scenario~\ref{scen:robot}, Fig.~\ref{fig:rob}), the robot
agent must be capable of finding out the duration of an episode of the
robot at the conveyor belt, \ie the
period length in time. This task also appears in speech and musical harmony
recognition, as illustrated in Fig.~\ref{fig:music}.
\end{scenario}

Since cognitive science may be defined as the study of the nature of intelligence and
thus of intelligent behavior, drawing on multiple disciplines, including
psychology, computer science, linguistics, and biology, we consider behavior and
periodicity analysis here, because it is obviously an important aspect of
intelligence. In particular, this holds for scenarios with several agents and/or
agents in dynamically changing environments, because it is the basis for
coordination and synchronization of (periodic) behavior of agents. For instance,
finding the way through a dynamic environment with many obstacles and crossing
traffic of a specific frequency, requires synchronization among agents, including
periodicity analysis.

One possible key for determining overall period lengths is auto-correlation, \ie
the cross-correlation of a signal with itself. It can be mathematically defined
by convolution (cf.~\cite{Ebe08}, see also Sect.~\ref{sec:fir}). However, we choose
another formalization here: We simply assume that a unit of a CTNN (cf.
Def.~\ref{def:unit}) can delay its incoming signals for a specific time delay
$\delta$. Then, a comparison of the original signal with the delayed one yields
the appropriate result. Eventually, biological neural networks, \eg the hearing
system in the brain, seem to be able to delay signals \cite{Lan97,Lan07}. But before we
present the CTNN model in more detail (Sect.~\ref{sec:cnn}), let us first discuss
related works that are more or less suitable for modeling the scenarios
introduced here.

\section{Neural Networks, Hybrid Automata, and Continuous Time}\label{sec:works}

\begin{figure}
	\centering
	\includegraphics[width=0.65\textwidth]{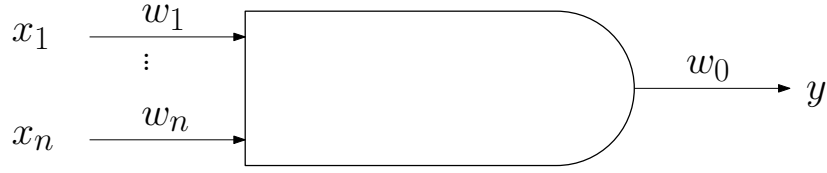}
	\caption{A unit of a neural network (scheme).}\label{fig:unit}
\end{figure}

The underlying idea that the original model of artificial neural networks tries
to capture is that the response function of a neuron is a weighted sum of its
inputs, filtered through a non-linear, in most cases sigmoidal function $h$:
\[ y = h(\sum_{i=1}^n w_i x_i) \]
$h$ is called activation function in this context. Often the logistic function,
defined by $x \mapsto 1/(1-e^{-x})$, is chosen.
Fig.~\ref{fig:unit} shows the general scheme of a unit of a neural
network with the inputs $x_1,\dots,x_n$ and one output $y$. Each incoming and
also the outgoing edge is annotated with a weight $w$.

\subsection{Fourier Neural Networks}

An obvious paradigm to combine neural networks with periodic input are so-called
\emph{Fourier neural networks} \cite{MA+04,Sil99}. They allow a more realistic
representation of the environment by considering input oscillation for
implementing and/or learning discrete functions or behavior. From a
neurophysiological point of view, they appear to be closer to reality, because
they model the signals exchanged between neurons as oscillations, making the
model to better agree with discoveries made in neurobiology. In \cite{Sil99},
the output function of a neuron is defined as $f(X) = \int_D c(X)\,
\varphi(X,Y)\,dY$, where $\varphi(X,Y)$ is some characteristics of the input
$X$, weighted by the coefficients $c(X)$, \ie, we get a weighted integral (replacing
the sum from above) of the inputs and their
characteristics. However for the computation, a discretized model given by the
equation $f^d(x_1,\dots,x_n) = h \left( \sum_{i} c_{i} \Pi_{j=1}^{n} \cos
(\omega_{ij}x_j+\varphi_{ij}) \right)$ is used with the sigmoidal logistic function $h$
from above in order to obtain output values in the interval $[0;1]$.

In \cite{MA+04}, Fourier neural networks with sinusoidal activation function
$h(x) = c\sin(ax+b)$ are considered. Additional non-linear (sigmoidal)
activation functions are not needed to express arbitrary functions in this case. In fact, the
sine function has the characteristics of a sigmoidal function in the interval
$[-\pi;\pi]$. All logical operators with two inputs (Scenario~\ref{scen:logic}) can be
implemented in this framework (see Fig.~\ref{fig:tab}) by only \emph{one} single unit
with sinusoidal activation function in contrast to the standard neural networks
with other, monotonously increasing activation functions. Even the odd parity function,
that returns $1$ iff an odd number of inputs is $1$, can be realized by only one
such unit, which is impossible for units with plain sigmoidal activation functions. 
However, learning these
neural networks is a difficult task, because sinusoidal activation functions are
non-monotonous. In addition, continuous time is not modeled explicitly in this
approach.

\begin{figure}
\begin{tabular}{*{5}{c@{\quad}}l}
function & \#\,inputs & $a$ & $b$ & $c$ & meaning\\[\medskipamount]
\textsf{AND} & $2$ & $\frac{\pi}{4}$ & $-\frac{\pi}{4}$ & $\sqrt{2}$ & logical conjunction\\
\textsf{XOR} & $2$ & $\frac{\pi}{2}$ & $-\frac{\pi}{2}$ & $1$ & exclusive or\\
\textsf{ODD} & $n$ & $\frac{\pi}{2}$ & $(n-1)\frac{\pi}{2}$ & $1$ & odd parity
\end{tabular}
\caption{Implementing logical functions for one Fourier neural network unit with
activation function $c\sin(ax+b)$. The Boolean values \emph{true} and
\emph{false} are represented by $+1$ and $-1$, respectively.}
\label{fig:tab} 
\end{figure}

\subsection{Continuous Neural Networks}

\cite{LB07} introduces neural networks with an uncountable number of hidden
units. While such a network has the same number of parameters as an ordinary
neural network, its internal structure suggests that it can represent some
smooth functions more compactly. \cite{LB07} presents another approach for
neural networks with an uncountable number of units, where the weighted
summation of input values is replaced by integration. Because of this, they are
called continuous neural networks. However, continuous time and hence temporal
processing is not modeled explicitly there, which is the primary goal of this
paper.

In \cite{LZ08}, specific neural networks are used in a non-linear system
identification algorithm for a class of non-linear systems. The algorithm
consists of two stages, namely preprocessing the system input and output and
neural network parameter estimation. However, first and foremost, it is only
applicable to the analysis of control systems with a special structure.

\subsection{Time-Delay Neural Networks}

A \emph{time-delay neural network} (TDNN) is a feed-forward multi-layer network that
contains two-dimensional layers (\eg spectrograms) which are sparsely connected
to each other and whose dimensions are defined by time and the number of features
\cite{LangWH90}. Here, each neuron is connected to a certain amount of consecutive
time frames in the lower layer. Shifted by one time unit, this receptive field
is then used to calculate the next time frame, keeping the same set of weights.
Therefore, a TDNN significantly reduces the amount of weights needed to be stored
and therefore minimizes the memory requirements of such processes as speech
recognition. However, a TDNN utilizes discrete time steps rather than continuous
time periods.

\subsection{Finite Impulse Response Perceptrons}\label{sec:fir}

Temporal processing in neural networks means to deal with dynamic effects and to
introduce time delays in the network structure \cite{Hay94}. Therefore, in the
\emph{finite-duration impulse response (FIR) model}, temporal processing is realized by a
linear, time-invariant filter for the synapse $i$ of a neuron $j$. Its impulse
response $h_{ji}(t)$ depends on a unit impulse at time $t=0$. Typically, each
synapse in the FIR model is causal and has a finite memory, \ie, $h_{ji}(t) = 0$
for $t < 0$ or $t > \tau$, with the memory span $\tau$ for all synapses. The response
of a synapse can be defined as the convolution (auto-correlation) of its impulse response with the
input $x_i(t)$. Thus, we can express the output as
$h_{ij}(t) \ast x_i(t) = \int_{-\infty}^{t} h_{ji}(u)x_i(t-u)du $.
The network activation potential over all $p$ synapses, with threshold $\theta_j$, is given by $v_j(t) = \left(
\sum_{i=1}^p \int_{0}^{\tau} h_{ji}(u)\,x_i(t-u)\,du \right) -
\theta_j$, where the overall output is the sigmoidal non-linear logistic
activation function applied to $v_j(t)$. With this, an artificial neuron can represent
temporal behavior. The FIR multi-layer perceptron, with its hidden and output
neurons based on this FIR model, has been applied for adaptive control, dynamic
system identification, and noise cancellation. Once trained, all synaptic
weights are fixed. Then, the network can operate in real time.

For computational reasons, FIR models approximate the continuous time with discrete
time steps. Hence, they are not as accurate as the CTNNs that we
will introduce in Section~\ref{sec:cnn}. Moreover and unlike a FIR model, a CTNN may
but does not have to be a recurrent network.

Instead of the
FIR model, where time is simulated by additional copies of a neuron for
different times (cf. Sect.~\ref{sec:scen}, Scenario~\ref{scen:robot}),
\emph{real-time recurrent networks} (cf.~\cite{Hay94}) are designed by using a
common neural model, where the temporal processing is realized by the feedback
of the network.

\subsection{Spiking Neural Networks}
A popular approach to simulating the electrophysical stimuli in the human brain
are \emph{spiking neural networks} \cite{GK02}. They build formal threshold models of
neuronal firing. Spikes are created when the membrane potential crosses some
threshold $\theta$. A well-known spiking neural network is the
\emph{integrate-and-fire model}, built up upon differential equations.
Here, an isolated neuron is stimulated by an external current. Then, the membrane
potential $u(t)$ reaches $\theta$ periodically and is reset immediately afterwards.
As an alternative model, the \emph{spike response model} (SRM) expresses the
output membrane as an integration over the spikes from the past.

Generally, if several neuron models are connected and combined to a network, the presynaptic
spikes generate a postsynaptic current pulse, where the firing time to the first
spike depends on the number of presynaptic spikes and its
amplitude depends on the membrane potential. The smaller it is, the higher the amplitude
of the input current.

\subsection{Hybrid Automata}\label{sec:ha}

\begin{figure}
	\centering
	\includegraphics[width=0.65\textwidth]{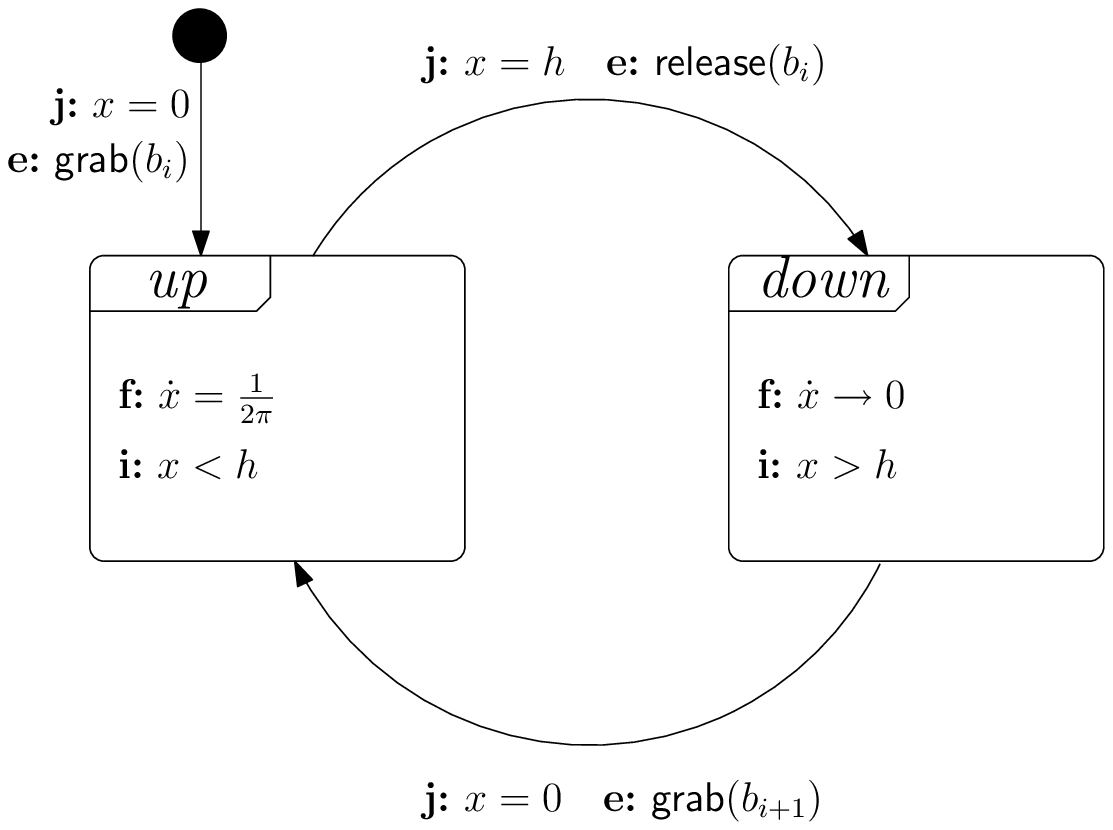}
	\caption{Hybrid automaton for the robot arm (Scenario~\ref{scen:robot}).
	Here, events are marked by \textbf{e}, flow conditions by \textbf{f},
	invariants by \textbf{i}, and jump conditions by \textbf{j}.}	
	\label{fig:hyb}
\end{figure}

Another model that allows to model discrete and dynamic changes of its
environment and hence continuous time are \emph{hybrid automata}, a
combination of Moore and Mealy automata \cite{Hen96}. A hybrid automaton is a
mathematical model for describing systems, where computational processes
interact with physical processes. In contrast to simple finite state automata,
well-known in computer science (see \eg \cite{Gil62,UML07}), their behavior is stated not only by discrete
state transitions, but also by continuous evolution. Hybrid automata consist of
a finite set of states and transitions between them. Thus, continuous flows
within states and discrete steps at the transitions are possible. If the state
invariants do not hold any longer, a discrete state change takes place, where a
jump condition indicates which transition shall be used. Then, a discrete step can be done,
before the next state is reached. States are annotated with invariants and flow
conditions, which may be differential equations. There, the continuous flow is
applied to the variables within the state invariants.  Thus, the behavior of the
robot in Scenario~\ref{scen:robot} can be described as shown in
Fig.~\ref{fig:hyb}. Hybrid automata, however, are not well-suited for mapping
continuous input with periodic behavior. In addition, (hybrid) automata
cannot be learned easily by examples as \eg neural networks.

\subsection{Central Pattern Generators}

For Scenario~\ref{scen:robot}, oscillating, periodic patterns must be generable.
This can be achieved, if a single unit is able to oscillate spontaneously, as we
will assume here (cf.~Def.~\ref{def:unit} below). Alternatively, recurrently connected
units can trigger each other, inducing periodic patterns. Such a system is called
\emph{central pattern generator} (CPG). They can be defined as neural networks that can
endogenously (i.e. without rhythmic sensory or central input) produce rhythmic
patterned outputs \cite{Hoo99} or as neural circuits that generate periodic
motor commands for rhythmic movements such as locomotion \cite{Kuo02}. CPGs have
been shown to produce rhythmic outputs resembling normal rhythmic motor pattern
production even in isolation from motor and sensory feedback from limbs and
other muscle targets. To be classified as a rhythmic generator, a CPG requires
two or more processes that interact such that each process sequentially
increases and decreases, and that, as a result of this interaction, the
system repeatedly returns to its starting condition.

The implementation of CPGs requires complex recurrent networks, which are
difficult wrt. specification and learning. Therefore, we will restrict attention
to feed-forward networks with oscillation units, which we will describe now.

\section{Continuous-Time Neural Networks}\label{sec:cnn}

We will now define \emph{continuous-time neural networks} (CTNN). With them, we
are capable of modeling the three general scenarios, introduced in
Sect.~\ref{sec:scen}. Even periodic behavior can be implemented easily by simple
non-recurrent networks, namely by means of so-called oscillating (sub-)units.
At first glance, CTNNs are very similar to standard neural
networks, because they also consist of an interconnected group of units. In
fact, a CTNN degenerates to an ordinary neural network, if the extended
functionality is not used. We distinguish several types of units (see Def.
\ref{def:basic}~and~\ref{def:unit}). 

\begin{definition}[input and output units, on-neurons]\label{def:basic}
In a CTNN, there may be one or more \emph{input and output units}. Input units do
not have any incoming edges, while output units do not have any outgoing edges.
In the following, we restrict our attention to networks with only one output unit.
The values of the input units $x_1(t),\dots,x_n(t)$ and of the output unit $y(t)$
depend on the time $t$. There may also be so-called \emph{on-neurons}, \ie units
without incoming edges, yielding a constant output $c$, independent from the
actual time $t$. As in ordinary neural networks, they are useful for defining
thresholds for different activation levels.
\end{definition}

In our model, as in standard neural networks, we assume that the input value of
a unit $j$ is a weighted sum of the incoming values, and we have a non-linear
activation function. But in addition, we have two further optional components
in each unit (for integration over time and for enabling oscillation) that may
be switched on or off. Furthermore, inputs may be delayed or not. This is
summarized in the following definition, leading to a unit with up to four
stages, called \emph{sub-units} in the sequel:

\begin{definition}[continuous neural network unit]\label{def:unit}
In general, a CTNN unit computes its output value $y(t)$ from its input values
$x_1(t),\dots,x_n(t)$, which may be the overall input values of the network or
the output values of immediate predecessor units, in four steps.
Each step yields the value $y_k(t)$ with $1 \le k \le 4$, where $f(t) = y_4(t)$.
For $k \ge 2$, the respective sub-unit may be switched off, \ie,
$y_k(t) = y_{k-1}(t)$ in this case. The four sub-units are:
\begin{enumerate}
  \item \textbf{summation:} The input value of the unit is the sum of the
	incoming values $x_i(t)$ with $1 \le i \le n$, each weighted with a
	factor $w_i$ and possibly delayed by a time amount $\delta_i$, which is $0$ by
	default:
	\[ y_1(t) = \sum_{i=1}^n w_i \cdot x_i(t-\delta_i) \] 
  \item \textbf{integration:} In certain cases, the integrated activity, \ie the
	average signal power, is useful. Therefore, we introduce an optional
	integration process, which is switched off by default.
	\[ y_2(t) = \sqrt{\frac{1}{\tau} \int\limits_{t-\tau}^t y_1(u)^2\,du} \]
	Note that, for $\tau \to 0$, we have $y_2(t) = |y_1(t)|$, \ie, the unit
	is switched off for positive values. If it is switched on,
	we take $\tau \to \infty$ by default. Alternatively, the statistical
	variance of $y_1(t)$ could be used here.
  \item \textbf{activation:} In order to be able to express general, non-linear
	functions, we need a non-linear activation function (cf.~\cite{Hay94}).
	Instead of the often used logistic function (cf.~Sect.~\ref{sec:works}), 
	we use the hyperbolic tangent here, because $\tanh(x) \approx x$ for
	small $x$ and the range of the hyperbolic tangent is $[-1;+1]$, which
	corresponds well to the range of sinusoidal periodic functions. We define:
	\[ y_3(t) = \frac{\tanh(\alpha \cdot y_2(t))}{\alpha} \]
	We make use of a factor $\alpha$ that retains these properties here.
	By default, $\alpha=1$. For $\alpha \to 0$, the sub-unit is switched off.
  \item \textbf{oscillation:} The unit can start to oscillate with a
	fixed (angular) frequency~$\omega$:
	\[ y_4(t) = y_3(t) \cdot \cos(\omega\,t) \]
	This corresponds to amplitude modulation of the input signal. In principle,
	other types of modulation, \eg frequency or phase modulation, would be
	possible (not considered here). For $\omega = 0$, this sub-unit is switched off.
\end{enumerate}
\end{definition}


With this type of units, all scenarios, introduced in Sect.~\ref{sec:scen}, can
be implemented. If the integration and the oscillation sub-unit is switched off,
the functionality of the unit is identical with that of standard neural network
units (cf. Sect.~\ref{sec:works} and \cite{Hay94,RN10}). Hence, all logical
Boolean functions (Scenario~\ref {scen:logic}) can be expressed easily, of
course, in contrast to Fourier neural networks, generally with hidden units.
Everything that can be expressed by an ordinary neural network can be expressed
by a CTNN, because the former one is a special case of a CTNN.

\begin{figure}
	\centering
	\includegraphics[width=\textwidth]{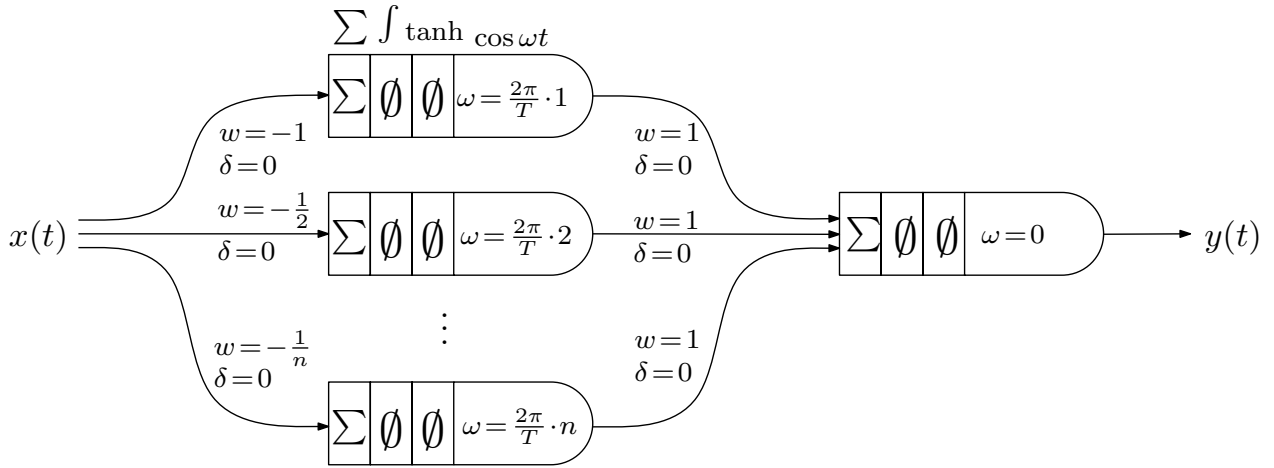}
	\caption{Network with several oscillating units for Scenario~\ref{scen:robot}.
	Sub-units, that are switched off, are marked with $\emptyset$.}\label{fig:network}
\end{figure}

Scenario~\ref{scen:robot} can be implemented with several oscillating units,
\ie $\omega_k \neq 0$, because it is known from the study of Fourier series, that
arbitrary periodic functions can be written as the sum of simple waves
represented by sines and cosines. For the sawtooth-like graph
(Fig.~\ref{fig:saw}), we have $f(x) = \frac{h}{2} - \frac{h}{\pi}
\sum_{k=1}^\infty \frac{1}{k} \cdot \sin(\frac{2\pi}{T}kx)$. The latter sum
may be approximated by the first $n$ summands, which can be expressed by $n$
oscillating CTNN units (see Fig.~\ref{fig:network}).

\begin{figure*}
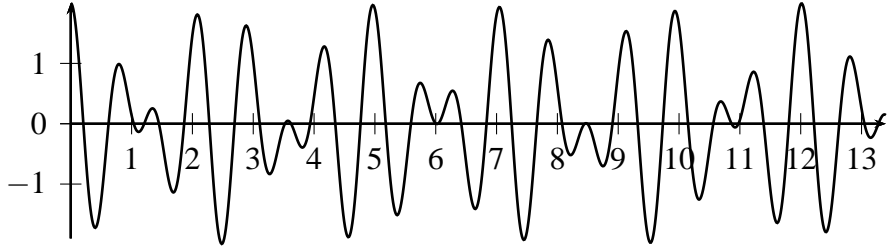

  \centering
  \graph{-1.9}{2}{cos(\dpi*x)+cos(\dpi*\rtw*x)}
  \caption{Complex periodic signal $x(t) = \cos(\omega_1t)+\cos(\omega_2t)$
	with $\omega_2/\omega_1 = \sqrt{2}$.}
  \label{fig:tritone}
\end{figure*}

In Scenario~\ref{scen:period}, we have to find out the period length $T$ of a
task automatically from a complex signal, \eg the duration of an episode of the
robot at the conveyor belt (Scenario~\ref{scen:robot}, Fig.~\ref{fig:rob}). For this, consider the
function $x(t) = \cos(\omega_1t)+\cos(\omega_2t)$, whose overall period length
depends on the ratio $\omega_2/\omega_1$. Let $\omega_1 = 2\pi$ and $\omega_2 =
\sqrt{2}\omega_1$. The corresponding graph for $x(t)$ is shown in
Fig.~\ref{fig:tritone}. In order to determine the overall period length, we must
be able to find out the so-called missing fundamental frequency, \ie, we have to
find a time duration $T$ such that $x(t)-x(t-T)$ becomes zero. Applying the
least squares method, this could be turned in finding the minima (almost zeros)
of $1/T \int_0^T (x(u)-x(u-T))^2\,du$. Therefore, we overlap the original signal
($\delta=0$, $w=1$) with a phase-shifted and inverted copy of itself
($\delta=T$, $w=-1$), which yields an effect of \emph{comb filtering} (cf.~\cite{Har97,Lan07}).

\begin{figure*}
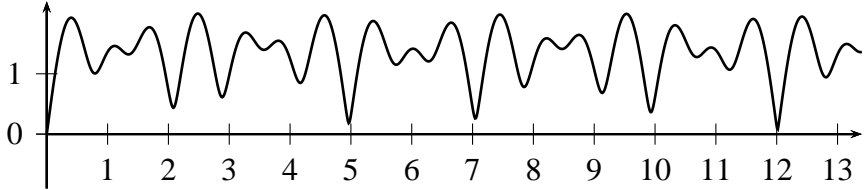

  \centering
  \graph{-0.9}{2.2}{sqrt(2-cos(\dpi*x)-cos(\dpi*\rtw*x))}
  \caption{Periodicity analysis for complex signal $x(t)$. The graph shows the
	output of the comb-filtered signal in dependency from the delay in time
	$T$, checking the period length of the overall signal, with main minima
	at $T=5$ and $T=12$. It is constant wrt. the current time $t$.}
  \label{fig:ana}
\end{figure*}

Fig.~\ref{fig:ana} shows the graph for the square root of the latter integral in dependency from
$T$, which can be achieved by switching on the integral sub-unit. It has minima
near $5$ and $12$ (and also near $7$ and $10$) which alternatively can be
derived by employing the so-called Stern-Brocot approximating the ratio
$\omega_2/\omega_1$ \cite{GKP94,Sto09,Sto15}. Thus, the
corresponding CTNN unit yields approximately constant output wrt. $t$, namely the
values shown in the graph in Fig.~\ref{fig:ana}, where small values near $0$
indicate periodicity. This procedure allows us to express analysis of periodic
behavior as desired. From a technical point of view, the integration unit leads
to \emph{phase locking}, \ie, more or less chaotic actions of individuals shift
to the ordered actions of the whole system, which is also required to detect
(periodic) patterns.

\subsection{Training Continuous-Time Neural Networks}

CTNNs are very expressive. The price for this, of course, is that training these
networks is a complex, yet feasible task. We will only sketch a learning
procedure here. We assume that the network has $n$ input units and only one
output unit. For the learning process, sample input and output values must be
known, \ie $x_i(t)$ and $y(t)$ for $1 \le i \le n$ and for sufficiently many
time points $t$. For each unit $i$ with output $a_i(t)$ and for each link between
the units $i$ and $j$, in principle, the parameters $\tau_i$, $\alpha_i$,
$\omega_i$, as well as $w_{i,j}$ and $\delta_{i,j}$ have to be determined. For
simple multi-layer feed-forward networks, only the link weights $w_{i,j}$ are
learned by the well-known backpropagation algorithm. It is a kind of gradient descent search in the error
space, \ie, the sum of squared errors is minimized. This procedure can be
enhanced and adapted to CTNNs. The error can be defined by \[ E = \frac{1}{2}
\sum_{k,l} \big(\hat{y}_l(t_k)-y_l(t_k)\big)^2 \] where $\hat{y}_l(t_k)$ denotes the
intended target value of output unit $l$ at time $t_k$. The values of all output
and hidden units $i$ can be computed recursively by the formula $a_j = f_j(t)$
(cf. Def.~\ref{def:unit}). The first derivatives wrt. all parameters mentioned
above yield the gradient for the search. A parameter $z$ can then be corrected
by the term $-\eta \cdot E'(z) \cdot a_j(t_k)$, where $\eta>0$ is the learning
rate and $E'(z)$ is the first derivative of $E$ wrt. $z$. Here output and hidden
units have to be distinguished (cf.~\cite{Hay94}). The implementation of
this procedure is subject of future work.

In this context, the time-dependent parameters $\delta$ and $\tau$ require a
special procedure: First, an output value $y(t)$ may depend on an input value
in the past, if some $\delta$ values are non-zero. Then, the corresponding input
value $x_i(t^*)$ with $t^*<t$ has to be interpolated, if the precise value is not
given. Second, if a $\delta$ value shall be learned, then the first derivative
of the respective input value $x_i(t-\delta)$ wrt. $\delta$ has to be estimated,
too. Finally, if some value of $\tau$ is not fixed to zero, then also integrals
have to be approximated.

\section{Conclusions}\label{sec:conc}

In this paper, we presented work on neural networks with continuous time.
These networks can support the modeling of behavior synthesis and analysis in robotics and for
cognitive systems. For arbitrary continuous, periodic input, the robot or the
agent in general has to react continuously and within a certain time interval.
Hence, complex, physical and/or cognitive processes can be modeled adequately by
a CTNN. A CTNN without recurrence and constant values for the angular frequencies
$\omega_k$ in the oscillation sub-units and switched-off integration sub-units
correspond to standard neural network units in principle. Thus, the classical
backpropagation method can be adapted for learning a CTNN from examples, where
a set of input and output values must be given for different time points $t$ (see also above).

\bibliographystyle{plain}
\bibliography{contnet}

\end{document}